\pdfoutput=1
\documentclass[11pt]{article}

\usepackage{ACL2023}

\usepackage{times}
\usepackage{latexsym}

\usepackage[T1]{fontenc}
\usepackage[utf8]{inputenc}

\usepackage{microtype}

\usepackage{inconsolata}

\usepackage{multirow}
\usepackage{multicol}

\usepackage{graphicx}
\usepackage{bm}
\usepackage{bbm}
\usepackage{amsmath}
\usepackage{amsfonts}
\usepackage{booktabs}
\usepackage{tikz}

\usepackage{makecell}
\usepackage{float}
\usepackage{subfigure}
\usepackage{color}
\usepackage{amssymb}
\usepackage{utfsym}
\usepackage{enumitem}
\usepackage{algpseudocode}
\usepackage{algorithm}
\usepackage{colortbl}
\usepackage{xcolor}

\definecolor{lightgray}{RGB}{227,227,227}
\definecolor{lightlightgray}{RGB}{240,240,240}

\title{Warmup-Distill: Bridge the Distribution Mismatch between Teacher and Student before Knowledge Distillation}

\author{
  Zengkui Sun\textsuperscript{1}\thanks{ \ \ Work was done when Zengkui Sun was an intern at Pattern Recognition Center, WeChat AI, Tencent Inc, China.},
  Yijin Liu\textsuperscript{2},
  Fandong Meng\textsuperscript{2},
  Yufeng Chen\textsuperscript{1}\thanks{ \ \ Yufeng Chen is the corresponding author.},
  Jinan Xu\textsuperscript{1},
  and Jie Zhou\textsuperscript{2} \\
  \textsuperscript{1}Beijing Key Laboratory of Traffic Data Mining and Embodied Intelligence, \\Beijing Jiaotong University, Beijing, China \\
  \textsuperscript{2}Pattern Recognition Center, WeChat AI, Tencent Inc, China \\
  \texttt{\{zengksun,jaxu,chenyf\}@bjtu.edu.cn} \\
  \texttt{\{yijinliu,fandongmeng,withtomzhou\}@tencent.com} \\
}

\begin{document}
\maketitle
\begin{abstract}
The widespread deployment of Large Language Models (LLMs) is hindered by the high computational demands, making knowledge distillation (KD) crucial for developing compact smaller ones. 
However, the conventional KD methods endure the distribution mismatch issue between the teacher and student models, leading to the poor performance of distillation.
For instance, the widely-used KL-based methods suffer the mode-averaging and mode-collapsing problems, since the mismatched probabitliy distribution between both models.
Previous studies mainly optimize this issue via different distance calculations towards the distribution of both models.
Unfortunately, the distribution mismatch issue still exists in the early stage of the distillation.
Hence, to reduce the impact of distribution mismatch, we propose a simple yet efficient method, named \textbf{Warmup-Distill}, which aligns the distillation of the student to that of the teacher in advance of distillation.
Specifically, we first detect the distribution of the student model in practical scenarios with its internal knowledge, and then modify the knowledge with low probability via the teacher as the checker.
Consequently, Warmup-Distill aligns the internal student's knowledge to that of the teacher, which expands the distribution of the student with the teacher's, and assists the student model to learn better in the subsequent distillation.
Experiments on the seven benchmarks demonstrate that Warmup-Distill could provide a warmup student more suitable for distillation, which outperforms the vanilla student by as least +0.4 averaged score among all benchmarks.
Noteably, with the assistance of Warmup-Distill, the distillation on the math task could yield a further improvement, at most +1.9\% accuracy.

\end{abstract}

\begin{figure}[t!]
\begin{center}
    % \scalebox{0.475}{
    \resizebox{0.475\textwidth}{!}{
        \includegraphics[width=1\textwidth]{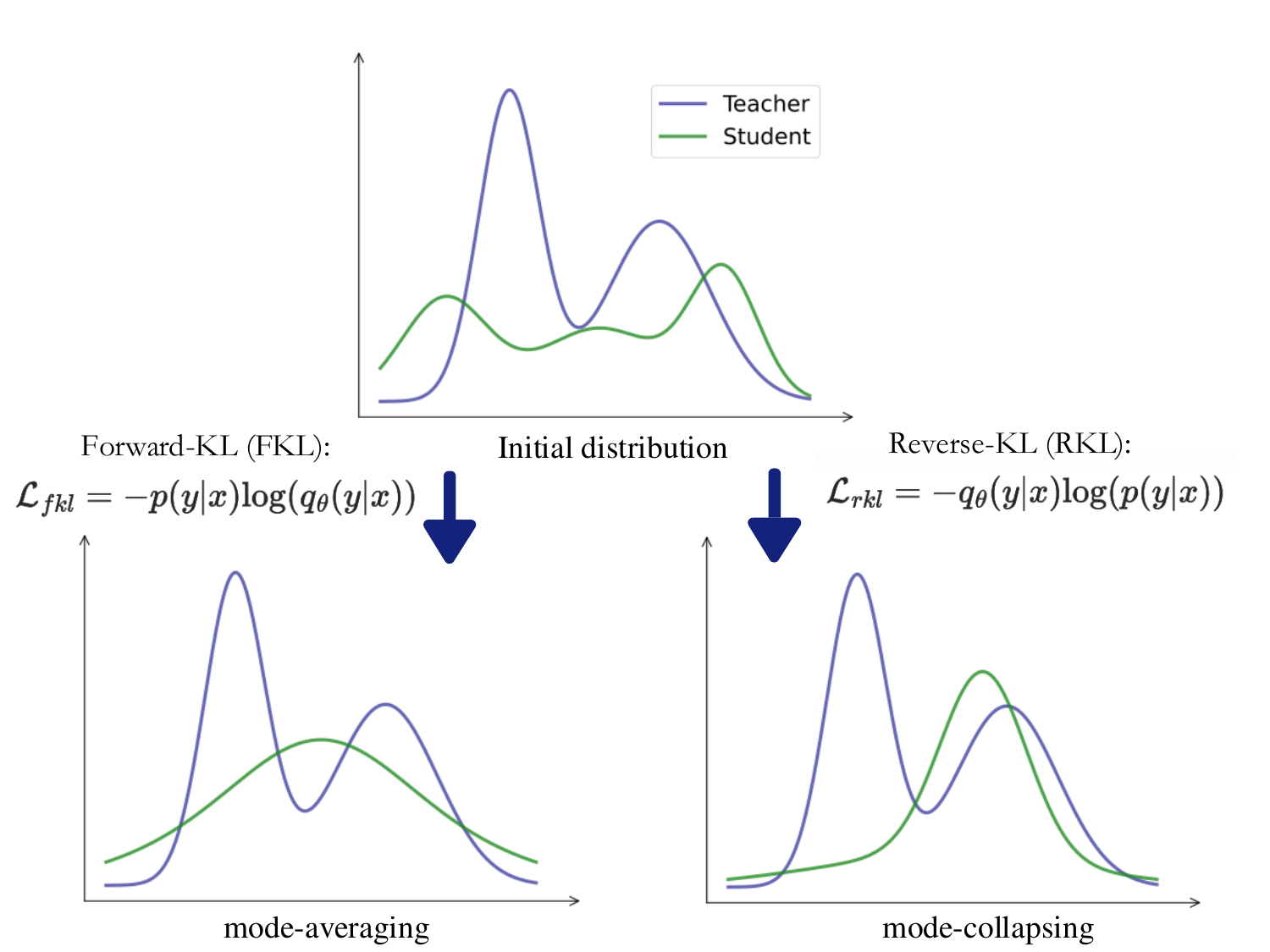}
    } 
    \caption{
        The Distribution Mismatch issue during distillation.
        Blue curve: teacher distribution $p(y|x)$;
        Green curve: student distribution $q_{\theta}(y|x)$.
    } 
    \label{fig: distribution_mismatch}  
\vspace{-6pt}
\end{center} 
\end{figure}

\section{Introduction}
Recently, Large Language Models (LLMs) have demonstrated remarkable capabilities in complex tasks \citep{brown2020language, ouyang2022training, wang2023chatgpt, Openo1, deepseekr1lite}.
However, the substantial computational overhead and memory consumption inherent in LLMs during inference present significant challenges for practical deployment \citep{agarwal2023gkd, xu2024speculative}.
Therefore, compressing the large models to the smaller, faster models while maintaining their performance is crucial \citep{buciluǎ2006model}.

Knowledge Distillation (KD) \citep{hinton2015distilling} is a widely used technique to compress LLMs by transferring knowledge from a larger teacher model to a smaller student model.
Conventional KD approaches fall into two frameworks, \emph{i.e.}, black-box KD and white-box KD.
Black-box KD adopts the high-quality teacher-generated output as the one-target label to optimize the student \citep{kim2016sequence, fu2023specializing}.
By contrast, white-box KD usually minimizes the distance (\emph{e.g.}, KL divergence) between the output distributions of the teacher and student \citep{hinton2015distilling, gu2024minillm, ko2024distillm}.
However, these above KD methods suffer the distribution mismatch issue \citep{agarwal2023gkd, wen2023f, ko2024distillm}.
Displayed as Figure \ref{fig: distribution_mismatch}, the KL-based methods endure the mode-averaging or mode-collapsing issue, where the student is misleaded by the distribution mismatch \citep{wen2023f, ko2024distillm}.
% 
% To reduce the impact of this issue, previous studies  the function.
Previous studies mainly optimize this issue via different distance calculations towards the distribution of both models \citep{wen2023f, gu2024minillm, ko2024distillm}.
However, the distribution mismatch issue still exists in the early stage of the distillation.

In this work, we introduce \textbf{Warmup-Distill} that firstly extends the distribution of the student by the teacher as the warmup stage, then conduct the distillation to optimize the student model with less distribution mismatch.
In the warmup stage, Warmup-Distill utilizes the student-generated samples to present the distribution of the student within its internal knowledge.
Consequently, Warmup-Distill selectively filters out low-probability intermediate tokens that exhibit low generation likelihood in the teacher model, subsequently performing resampling operations directly from the teacher's probability distribution.
% 
% In this way, the student acquires the teacher's knowledge distribution while maintaining semantic coherence with the teacher's expertise.
In this way, Warmup-Distill acquires the knowledge with the mismatch issue between both models in the practical scenario, before the distillation.
Further, Warmup-Distill reduces the degree of distribution mismatch by optimizing the knowledge within the student model from its distribution to the teacher.
Overall, the warmup stage could bridge the mismatched distribution between the teacher and student model, and prepare a student more suitable for distillation.

We evaluate Warmup-Distill on the instruction-following and math benchmarks with two different model architectures, \emph{i.e.}, Qwen \citep{qwen2.5} and Llama3 \citep{meta2024introducing} series.
Experimental results showcase that the Warmup-Distill could effectively reduce the interference of distribution mismatch to the performance of distillation.
With the warmup student prepared by Warmup-Distill, the performance of distillation could obtain a further improvement on the current widely used KD techniques, including the black-box and white box KD.
Specifically, on two task types and seven benchmarks, the KD techniques could achieve +0.4 average score improvement, and could yield at most +1.9\% accuracy improvement on the math task with the warmup student.

To sum up, the contributions are as follows\footnote{Codes are released at \url{https://github.com/Acerkoo/WarmupDistill}.}:
\begin{itemize}[leftmargin=*,topsep=0pt]
\setlength{\itemsep}{0pt}
\setlength{\parsep}{0pt}
\setlength{\parskip}{0pt}
    \item We propose a new perspective to mitigate the distribution mismatch issue during distillation, \emph{i.e.}, \textbf{Warmup-Distill}, which captures and reduces this issue before distillation.
    % 
    % \item We present the annotation-free approach to construct preference pairwise pairs to align the student with the teacher models, and prepare the aligned student to be distilled from the teacher.
    \item In Warmup-Distiil, we present a simple yet effective solution to capture and align the mismatched distribution of the student to the teacher in the practical scenario.
    \item Experiments demonstrate that our proposed Warmup-Distill could effectively reduce the interference of distribution mismatch during distillaiton, and take a further improvement to the performance of the current KD techniques.
\end{itemize}

\begin{figure*}[t]
\begin{center}
    % \scalebox{0.475}{
    \resizebox{0.9\textwidth}{!}{
        \includegraphics[width=1\textwidth]{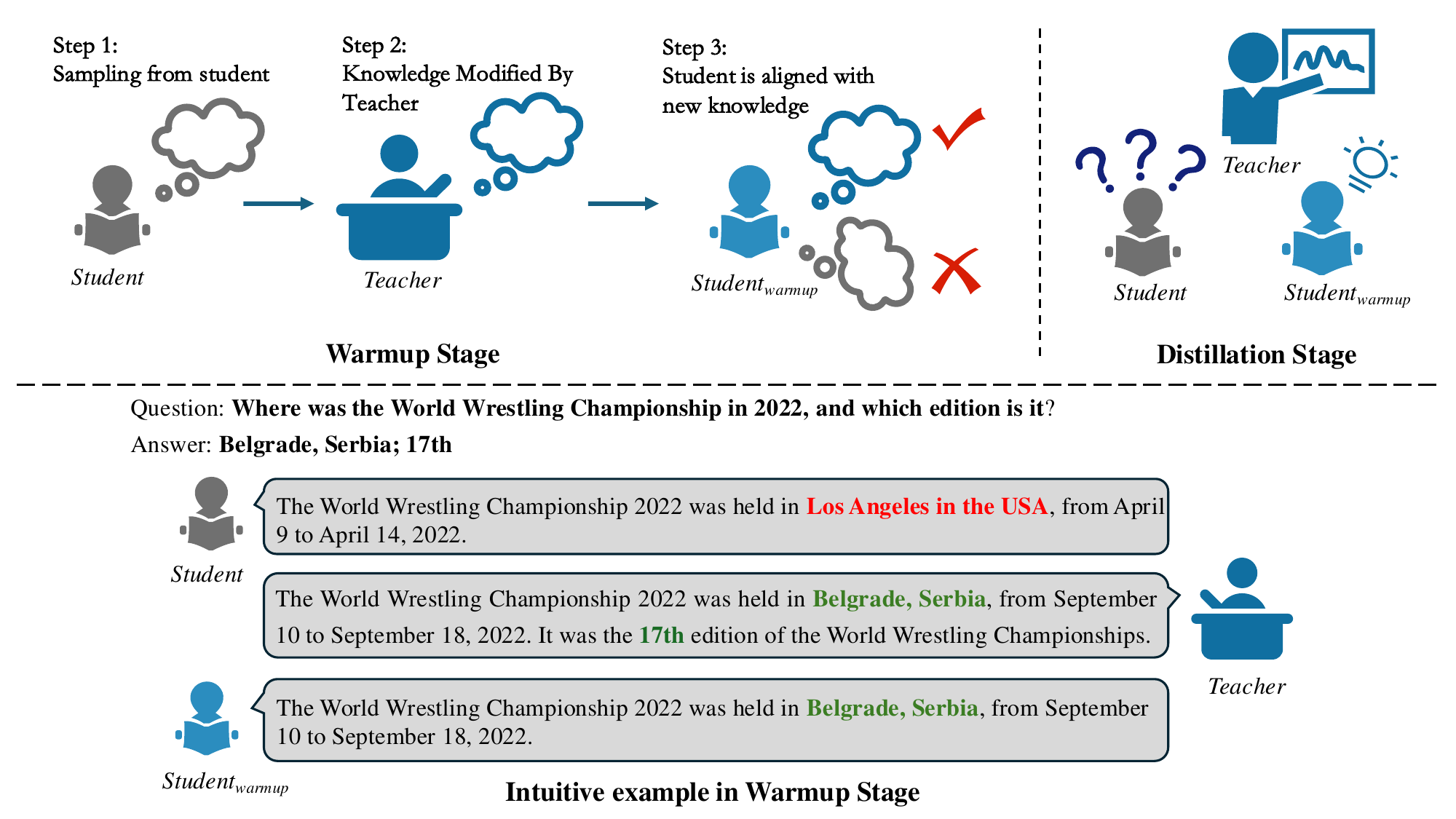}
    } 
    \caption{
        Illustration of \textbf{Warmup-Distill}. 
        Warmup-Distill mainly focuses on the warmup stage, where the distribution of the student model is sampled and aligned to the teacher's one, to reduce the distribution mismatch issue for the subsequent distillation.
        % 
        % Before the distillation, the d between the teacher and student models is initially filled.
    } 
    \label{fig: method}  
\vspace{-6pt}
\end{center} 
\end{figure*}

\section{Background}
% In this section, we first review the widely-used KD techniques, 
% \subsection{Knowledge Distillation}
Given a sequence $\mathbf{x}$, current LLMs generally learn the objective of casual language modeling through cross-entropy loss:
\begin{equation} \label{eq: ce_loss}
    \mathcal{L}_{ce} = - \sum_{i}^{|\mathbf{x}|} \log q_{\theta}(x_i|\mathrm{x}_{<i}), 
\end{equation}
where $q_{\theta}(x_i|\mathrm{x}_{<i})$ denotes the probability of the student model to the $i$-th target token $x_i$ conditioning on the context $\mathrm{x}_{<i}$.
In practice, the black-box KD methods, \emph{e.g.}, SeqKD \citep{kim2016sequence}, leverage this objective to optimize the student model on the high-quality teacher-generated data.

Compared to black-box KD methods, the white-box KD methods first feed this sequence into the teacher model to obtain its token-level probability distributions $p(x_i|\mathbf{x}_{<i})$.
Then, applying the following loss to minimize the distance between the student distribution $q_{\theta}(x_i|\mathbf{x}_{<i})$ and the teacher distribution $p(x_i|\mathbf{x}_{<i})$:
\begin{equation} \label{eq: kd_loss}
\small 
\begin{split}
    \mathcal{L}_{kd}&=\sum_i \mathcal{D}(p(x_i|\mathbf{x}_{<i};\tau) || q_{\theta}(x_i|\mathbf{x}_{<i};\tau)) \\
                    &=-\sum_i p(x_i|\mathbf{x}_{<i};\tau) \mathrm{log}(q_{\theta}(x_i|\mathbf{x}_{<i};\tau)),
\end{split}
\end{equation}
where $\mathcal{D}(\cdot || \cdot)$ is the distance function that measures the distance between the two distributions (\emph{e.g.}, KL divergence) and $\tau$ is the temperature coefficient to control the sharpness of the distributions. 

As previous studies \citep{wen2023f, ko2024distillm, xu2024speculative} stated, the KL-based losses suffer the issue of distribution mismatching.
For instance, in Equation \ref{eq: kd_loss}, the loss term $-\mathrm{log} q_{\theta}(x_i|x_{<i})$ goes to infinity when the student assigns a low probability to the target token.
As a result, minimizing KL forces the student model to spread its probability mass widely over the vocabulary, and the learned distribution may not capture any mode of teacher distribution, \emph{i.e.}, the mode-averaging problem.
By contrast, in another widely used Reverse-KL \citep{tu2020engine} losses, which swaps the position $p(\cdot)$ and $q_{\theta}(\cdot)$, minimizing KL leads the student to learn the high-probability regions of the teacher model, \emph{i.e.}, the mode-collapsing problem.
Hence, aligning the distribution between both models is crucial for distillation.

\section{Methodology}
We introduce \textbf{Warmup-Distill}, a simple yet effective method to prepare the student model to suffer the less distribution-mismatch issue during distillation.
The core idea is to capture the knowledge with mismatched distribution between the teacher and student models, and then align the distribution of the student to the teacher's, \textbf{before} distillation.

As diagrammed in Figure \ref{fig: method}, Warmup-Distill consists of three main steps in the warmup stage:
\begin{itemize} [leftmargin=*,topsep=0pt]
\setlength{\itemsep}{0pt}
\setlength{\parsep}{0pt}
\setlength{\parskip}{0pt}
    \item Sampling the output from the student model as its internal distribution in practical scenarios;
    \item Compared the distribution between both models, and modify the student's output from the teacher's distribution (\S\ref{sec: pairwise});
    \item Aligning the internal distribution of student to the teacher’ one (\S\ref{sec: align}); 
\end{itemize}

\subsection{Distribution Comparison} \label{sec: pairwise}
After sampling the knowledge from the student model, we leverage the teacher model to detect its probability distribution on these sampled sequences, and then modify the tokens exhibiting low probability.

Given the student model $q_{\theta}$ and input $x$, we could sample $y\sim q_{\theta}(x)$, which represents the internal distribution within $q_{\theta}$ towards the input $x$ in the practical scenarios.
Assuming the sampled sequence as $y=\{y_1, y_2, \dots, y_{m}\}$ with $m$ tokens, the teacher model detects the sequence by its probability distribution:
\begin{equation}
\small
\begin{split}
    P(y | x) = \{p(y_i | x, y_{<i}), 1 \le i \le m\},
    % p(y) = \{ r_{p_\theta}(y^{(T)}_{i}) - r_{q_\theta}(y^{(T)}_{i}),  0 \le i \le m\}, 
\end{split}
% \label{eq: gap}
\end{equation}
where $P(y|x)$ denotes the sequence of probabilities of the teacher model $p$ towards each token in student-generated sequence $y$.
Similarly, we obtain the probability scores of the student model $q_{\theta}$ to the same sampled sequence:
\begin{equation}
\small
\begin{split}
    Q_{\theta}(y | x) = \{q_{\theta}(y_i | x, y_{<i}), 1 \le i \le m\}.
    % p(y) = \{ r_{p_\theta}(y^{(T)}_{i}) - r_{q_\theta}(y^{(T)}_{i}),  0 \le i \le m\}, 
\end{split}
% \label{eq: gap}
\end{equation}

The higher score in $P(y|x)$ denotes the corresponding token is more fitted to the probability distribution within the teacher, meaning that the few distribution-mismatch in the subsequent distillation.
Further, the larger margin between $P(y|x)$ and $Q_{\theta}(y|x)$ denotes the larger distribution mismatch between the teacher and student model, which is required to be captured.
The sequence of margins between scores could be represented as: 
\begin{equation}
\small
\begin{split}
\mathrm{M}(y|x) = \{q_{\theta}(y_i|x, y_{<i}) - p(y_i|x, y_{<i}),\\ 1\le i\le m\}.
\end{split}
\label{eq: margin}
\end{equation}

On this basic, we set the threshold $\eta$ to filter out the first token with a larger value than $\eta$ in $\mathrm{M}(y|x)$.
Besides, we pass the subsequent tokens in sequence to avoid the error accumulation and replace these tokens with the continually generated tokens by the teacher.
Since the continually generated tokens may be out of distribution of the teacher \citep{ko2024distillm}, we introduce a extra reward model to ensure the sequence with continually generated tokens is better than the original one for further optimization (details could be found in Table \ref{tab: data_stat}).

% To better understand, 
% Consequently, we filter the preference sequence $gap(y^{(T)})$ and obtain the in-distribution sequence $y^{(T)'}$, which drops the rest of the tokens after the first out-of-distribution token:
% Consequently, the teacher selectively filter out the tokens with the low probability score that is the distribution different part between both models:
% Consequently, to capture the distribution 
% \begin{equation}
% \begin{split}
%     pos(y) &= \mathop{\arg\min}\limits_{i} \{gap[i] \ge \tau,  0 \le i \le m\}, \\
%     y^{(T)'} &=\{y^{(T)}_0, y^{(T)}_1, \dots, y^{(T)}_{pos(y^{(T)})-1}\},
% \end{split}
% \label{eq: pos}
% \end{equation}
% where $\tau$ is the hyperparameter to control the degree of filter.

% Finally, the student finishes the in-distribution sequence $y^{(T)'}$ with its internal knowledge, denoted as $y^{(T)}_s$.
% 
% Hence, we could build the pairwise response data for the input $x$:
% \begin{equation}
%     (\hat{y}_+, \hat{y}_-) := (y^{(T)}, y^{(T)}_s)
% \end{equation}

% Corresponding, for the student-generated sequence $y^{(S)}=\{y^{(S)}_0, y^{(S)}_1, \dots, y^{(S)}_{m'-1}\}$ with $m'$ tokens, we leverage the teacher model to generate $y^{(S)}_t$ via Eq.\ref{eq: gap} and Eq.\ref{eq: pos} and conduct the following pairwise data:
% \begin{equation}
%     (\hat{y}_+, \hat{y}_-) := (y^{(S)}_t, y^{(S)})
% \end{equation}

\subsection{Distribution Alignment} \label{sec: align}
 To align the distribution of the student model to the teacher's before the distillation, we leverage the two types of generated data in \S\ref{sec: pairwise}, \emph{i.e.}, the original data sampled from student and the teacher continual generated.
 Specifically, we conduct pairwise data to guide the distribution of the student model from the original one to that of the teacher.
 % we conduct the Direct Preference Optimization (DPO) \citep{rafailov2024direct} to optimize the student with the pairwise data in \S\ref{sec: pairwise}.

% Given two comparable objects or events, a common model for comparing the probabilities of their selection is the Bradley-Terry (BT;~\citealp{Bradley1952RankAO}) model. We consider output $y_1$ and $y_2$, and the reward function that measures the gain of choosing an output is denoted $r(y)$. BT model can be used to measure the probability that we choose $y_1$ instead of $y_2$:
Assuming the original sequence as $y_{-}$ and teacher continual generated one as $y_{+}$, the pair of data is formulated as $(y_{+}, y_{-})$.
Further, we utilize the Direct Preference Optimization (DPO) \citep{rafailov2024direct}, which leverages the Bradley-Terry (BT;~\citep{bradley1952rank}) model to compare the probabilities of both sequences and align the preference of the student model from $y_{-}$ to $y_{+}$.
For the pair of $(y_+, y_-)$, the BT model can be used to measure the probability of both sequences as follows:
\begin{equation}
\begin{small}
\begin{aligned}
  \label{eq:example}
 Pr(y_+ \succ y_-) &=\frac{\exp (r(y_+))}{\exp (r(y_+))+\exp (r(y_-))}\\
                   &= \sigma[r(y_+)- r(y_-)] ,
\end{aligned}
\end{small}
\end{equation}
where $\sigma$ is the function $\sigma(x) = \frac{1}{1 + \exp(-x)}$, and $r(\cdot)$ denotes the reward model.

Consequently, DPO utilizes the BT model to maximize the probability of the preferred output $y_+$ and minimize that of the undesirable output $y_-$.
The optimization objective is formulated as:
\begin{equation}
\begin{small}
\begin{aligned}
\mathcal{L}_{dpo}(\theta) &= -\mathbb{E}_{_{(x,y_+,y_-) \sim D}} [\\
&\log \sigma (\beta \log \frac{\pi_{\theta}(y_+|x)}{\pi_{ref}(y_+|x)}  - \beta \log \frac{\pi_{\theta}(y_-|x)}{\pi_{ref}(y_-|x)})],
\end{aligned}
\end{small}
\end{equation}
where $\beta$ is a hyperparameter.

We leverage $\mathcal{L}_{dpo}$ as the loss function to optimize the distribution of the student model and obtain the warmup stduent$_{warmup}$ for the subsequent distillation.
% To take a fine-grained process supervision, we follow Step-DPO \citep{lai2024step} to identify detailed preference difference between the teacher and student.
% 
% For the pair of $(\hat{y}_+, \hat{y}_-)$, we mainly compare the probability of the parts with different tokens as follows:
% \begin{equation}
% \begin{small}
% \begin{aligned}
% \mathcal{L}_{dpo}(\theta) = -\mathbb{E}_{_{(x, y_c, \hat{y}'_+,\hat{y}'_-) \sim D}} [\log \sigma (\beta \log \frac{\pi_{\theta}(\hat{y}'_+|x, y_c)}{\pi_{ref}(\hat{y}_+|x, y_c)} \\
%                                     - \beta \log \frac{\pi_{\theta}(\hat{y}'_-|x)}{\pi_{ref}(\hat{y}'_-|x, y_c)})],
% \end{aligned}
% \end{small}
% \end{equation}
% % 
% where $y_c$ denotes the common prefix tokens between $\hat{y}_+$ and $\hat{y}_-$, while $\hat{y}'_+$ and $\hat{y}'_-$ represent the rest tokens of $\hat{y}_+$ and $\hat{y}_-$ without $y_c$, respectively.

\subsection{Posted Distillation}
The overall framework is illustrated in top part of Figure~\ref{fig: method}, and we post a case for a better understanding in the bottom part of Figure~\ref{fig: method}.
For the given question, the student could not generate the correct answer, \emph{i.e.}, the unexpected answer ``\texttt{Los Angeles in the USA}''.
Besides, the subsequent text has the accumulated error, \emph{i.e.}, \texttt{from April 9 to April 14, 2022}.
By capturing the probability margin, the teacher could correct this error, and further generate the correct answer for the second question ``\texttt{which edition is it}''.
After the distribution alignment, the student$_{warmup}$ could output the correct answer for the first question, with the less different knowledge distribution from the teacher.
With the student$_{warmup}$, Warmup-Distill then conducts the distillation via current KD techniques to transfer knowledge from teacher to student.
% 

% \begin{algorithm}[t]
%     \caption{\textbf{Post Distillation}} 
%     \label{algo: post_distill}
%     \small
% \end{algorithm}

\begin{table*}[t]
    \centering
    \resizebox{0.96\linewidth}{!}{
        \begin{tabular}{l|ccccc|cc|c}
            \toprule 
            \makecell[c]{\textbf{Methods}} & \textbf{Dolly} & \textbf{SelfInst} & \textbf{VicunaEval} & \textbf{S-NI} & \textbf{UnNI}  & \textbf{Math-500} & \textbf{GSM8K} & \textbf{Avg.($\Delta$)}\\
            % \bottomrule
            % \toprule
            \midrule
            \multicolumn{9}{c}{\textbf{Qwen2.5-7B-Instruct} $\rightarrow$ \textbf{Qwen2.5-1.5B-Instruct}} \\
            \midrule
            % \toprule
            % \hline
           % \rowcolor{lightgray} 
            \cellcolor{lightgray}{Teacher} & \cellcolor{lightgray}{30.61} & \cellcolor{lightgray}{24.16} & \cellcolor{lightgray}{22.58} & \cellcolor{lightgray}{42.97} &  \cellcolor{lightgray}{37.55} & \cellcolor{lightgray}{75.4} & \cellcolor{lightgray}{92.4} & \cellcolor{lightgray}{-} \\
            \midrule
            \cellcolor{lightlightgray}{Student} &\cellcolor{lightlightgray}{23.68} & \cellcolor{lightlightgray}{17.04} & \cellcolor{lightlightgray}{17.06} & \cellcolor{lightlightgray}{34.70} & \cellcolor{lightlightgray}{32.43} & \cellcolor{lightlightgray}{56.0} & \cellcolor{lightlightgray}{72.3} & \cellcolor{lightlightgray}{-} \\
            \quad + SeqKD & 27.14 & 21.85 & 19.98 & 40.57 & 35.82  & 57.9 & 75.4  & - \\
            \quad + SKD & 27.22 & 21.74 & 20.14 & 40.91 & 36.93 & 58.2 & 75.2 & - \\
            \quad + FKL & 28.29 & 22.17 & 21.69 & 42.85 & 38.84  & 57.2 & 75.1 & - \\
            \quad + RKL & 28.03 & 22.31 & 20.75 & 43.40 &  39.65  & 57.4 & 75.8 & - \\
            \quad + f-Distill & 27.82 & 22.83 & 20.38 & 42.78 & 38.38 & 57.6 & 74.9 & - \\
            \quad + s-FKL & 27.55 & 22.85 & 21.91 & 42.55 & 39.76 & 57.4 & 74.6 & - \\
            \quad + aKL  & 28.72 & 23.60 & 20.45 & 42.34 & 40.13 & 57.8 & 74.7 & - \\
            \midrule
            \cellcolor{lightlightgray}{Student$_{warmup}$ (Ours)} & \cellcolor{lightlightgray}{24.31}$_{+0.63}$ & \cellcolor{lightlightgray}{17.27}$_{+0.23}$ & \cellcolor{lightlightgray}{17.33}$_{+0.27}$ & \cellcolor{lightlightgray}{35.23}$_{+0.53}$ & \cellcolor{lightlightgray}{32.88}$_{+0.45}$ & \cellcolor{lightlightgray}{57.4}$_{+1.4}$ & \cellcolor{lightlightgray}{74.2}$_{+1.9}$ & \cellcolor{lightlightgray}{+0.77} \\
            % \midrule
            \quad + SeqKD & 27.91$_{+0.77}$ & 22.54$_{+0.69}$ & 20.21$_{+0.23}$ & 40.66$_{+0.09}$ & 37.48$_{+1.66}$ & 59.8$_{+1.9}$ & 76.3$_{+0.9}$ & +0.89 \\
            \quad + SKD & 28.04$_{+0.82}$ & 22.26$_{+0.52}$ & 20.64$_{+0.50}$ & 41.77$_{+0.86}$ & 38.20$_{+1.27}$ & 59.6$_{+1.4}$ & 75.9$_{+0.7}$ & +0.86 \\
            \quad + FKL & 28.81$_{+0.52}$ & 22.84$_{+0.67}$ & 21.53$_{-0.16}$ & 43.32$_{-0.16}$ & 38.98$_{+0.14}$ & 58.2$_{+1.0}$ & 75.4$_{+0.3}$ & +0.42\\
            \quad + RKL & 29.26$_{+1.23}$ & 22.93$_{+0.62}$ & 21.25$_{+0.50}$ & 43.79$_{+0.39}$ & 39.96$_{+0.31}$ & 58.4$_{+1.0}$ & 75.9$_{+0.1}$ & +0.59\\
            \quad + f-Distill & 28.16$_{+0.34}$ & 23.42$_{+0.59}$ & 21.07$_{+0.69}$ & 42.97$_{+0.19}$ & 38.74$_{+0.36}$ & 58.8$_{+1.2}$ & 76.0$_{+1.1}$ & +0.63\\
            \quad + s-FKL & 28.56$_{+1.01}$ & 23.47$_{+0.62}$ & 21.89$_{-0.02}$ & 42.60$_{+0.05}$ & 39.86$_{+0.10}$ & 58.6$_{+1.2}$ & 75.4$_{+0.8}$ & +0.53\\
            \quad + aKL & 29.69$_{+0.97}$ & 23.88$_{+0.28}$ & 20.59$_{+0.14}$ & 42.97$_{+0.63}$ & 39.95$_{-0.18}$ & 58.4$_{+0.6}$ & 75.7$_{+1.0}$ & +0.49\\
            \bottomrule
        \end{tabular}
    }
    \caption{
    The results of distillation from Qwen2.5-7B-Instruct to Qwen2.5-1.5B-Instruct. 
    Digital results of Instruct-following benchmarks (Dolly, SelfInst, VicunaEval, S-NI, and U-NI) represent the scores of Rouge-L, and the results of math benchmarks (MATH-500 and GSM8K) denote the accuracy.
    We split both types of results with `|'.
    The subscripts in the bottom part are the delta values between the distillation with the student$_{warmup}$ or vanilla student model, and we also report the averaged delta values in the last column.
    }
    \label{tab: main_results_qwen}
\end{table*}
    
\begin{table*}[t]
    \centering
    \resizebox{0.96\linewidth}{!}{
        \begin{tabular}{l|ccccc|cc|c}
            \toprule 
            \makecell[c]{\textbf{Methods}} & \textbf{Dolly} & \textbf{SelfInst} & \textbf{VicunaEval} & \textbf{S-NI} & \textbf{UnNI}  & \textbf{Math-500} & \textbf{GSM8K} & \textbf{Avg.($\Delta$)}\\
            % \bottomrule
            \midrule
            \multicolumn{9}{c}{\textbf{Llama3.1-8B-Instruct} $\rightarrow$ \textbf{Llama3.2-1B-Instruct}} \\
            \midrule
            % \toprule
            % \hline
            \cellcolor{lightgray}{Teacher} & \cellcolor{lightgray}{26.24} & \cellcolor{lightgray}{17.26} & \cellcolor{lightgray}{22.56} & \cellcolor{lightgray}{29.69} & \cellcolor{lightgray}{34.77}  & \cellcolor{lightgray}{51.4} & \cellcolor{lightgray}{83.5} & \cellcolor{lightgray}{-} \\
            \midrule
            \cellcolor{lightlightgray}{Student} &\cellcolor{lightlightgray}{24.32} & \cellcolor{lightlightgray}{10.52} & \cellcolor{lightlightgray}{19.09} & \cellcolor{lightlightgray}{16.88} &  \cellcolor{lightlightgray}{17.55} & \cellcolor{lightlightgray}{28.4} & \cellcolor{lightlightgray}{37.3} & \cellcolor{lightlightgray}{-} \\
            \quad + SeqKD & 24.44 & 11.99 & 20.55 & 19.09 &  19.92  & 29.0 & 42.8 & - \\
            \quad + SKD & 24.53 & 12.24 & 20.44 & 19.54 & 20.43 & 28.8 & 42.3 & - \\
            \quad + FKL & 24.20 & 12.65 & 22.53 & 20.06 & 20.91 & 29.2 & 42.8  & - \\
            \quad + RKL & 25.34 & 12.39 & 23.18 & 22.32 & 22.84 & 28.8 & 43.5 & - \\
            \quad + f-Distill & 23.91 & 11.80 & 22.68 & 19.59 & 19.96 & 28.4 & 42.7 & - \\
            \quad + s-FKL & 23.93 & 12.22 & 22.24 & 19.77 & 20.82 & 28.6 & 43.1 & - \\
            \quad + aKL & 24.14 & 12.08 & 22.90 & 22.05 & 22.41 & 29.2 & 43.7 & - \\ 
            \midrule
            \cellcolor{lightlightgray}{Student$_{warmup}$ (Ours)} & \cellcolor{lightlightgray}{24.43}$_{+0.11}$ & \cellcolor{lightlightgray}{10.92}$_{+0.40}$ & \cellcolor{lightlightgray}{19.74}$_{+0.65}$ & \cellcolor{lightlightgray}{17.41}$_{+0.53}$ & \cellcolor{lightlightgray}{17.61}$_{+0.06}$ & \cellcolor{lightlightgray}{28.8}$_{+0.4}$ & \cellcolor{lightlightgray}{41.1}$_{+3.8}$ & \cellcolor{lightlightgray}{+0.85} \\
            % \midrule
            \quad + SeqKD & 24.76$_{+0.32}$ & 12.37$_{+0.38}$ & 20.73$_{+0.18}$ & 19.15$_{+0.06}$ & 20.01$_{+0.09}$ & 30.6$_{+1.6}$ & 43.7$_{+0.9}$ & +0.51 \\
            \quad + SKD & 24.97$_{+0.44}$ & 12.84$_{+0.60}$ & 20.35$_{-0.09}$ & 19.72$_{+0.18}$ & 20.87$_{+0.44}$ & 30.0$_{+1.2}$ & 43.7$_{+1.8}$ & +0.64\\
            \quad + FKL & 24.52$_{+0.32}$ & 12.98$_{+0.33}$ & 22.75$_{+0.22}$ & 20.47$_{+0.41}$ & 20.78$_{-0.13}$ & 29.8$_{+0.6}$ & 44.1$_{+1.1}$ & +0.41 \\
            \quad + RKL & 25.77$_{+0.43}$ & 12.32$_{-0.07}$ & 23.74$_{+0.56}$ & 22.70$_{+0.38}$ & 22.95$_{+0.11}$ & 30.4$_{+1.6}$ & 43.9$_{+0.7}$ & +0.53 \\
            \quad + f-Distill & 24.04$_{+0.13}$ & 12.18$_{+0.38}$ & 22.61$_{-0.07}$ & 20.03$_{+0.44}$ & 20.34$_{+0.38}$ & 29.2$_{+0.8}$ & 44.2$_{+0.9}$ & +0.43 \\
            \quad + s-FKL & 24.35$_{+0.42}$ & 12.83$_{+0.61}$ & 22.65$_{+0.41}$ & 20.29$_{+0.52}$ & 21.35$_{+0.53}$ & 29.6$_{+1.0}$ & 43.8$_{+0.7}$ & +0.59 \\
            \quad + aKL & 24.46$_{+0.32}$ & 12.79$_{+0.71}$ & 23.32$_{+0.42}$ & 22.43$_{+0.38}$ & 22.38$_{-0.03}$ & 29.8$_{+0.6}$ & 44.1$_{+0.4}$ & +0.40 \\ 
            \bottomrule
        \end{tabular}
    }
    \caption{
    The results of distillation from Llama3.1-8B-Instruct to Llama3.2-1B-Instruct. 
    Digital results of Instruct-following benchmarks (Dolly, SelfInst, VicunaEval, S-NI, and U-NI) represent the scores of Rouge-L, and the results of math benchmarks (MATH-500 and GSM8K) denote the accuracy.
    We split both types of results with `|'.
    The subscripts in the bottom part are the delta values between the distillation with the student$_{warmup}$ or vanilla student model, and we also report the averaged delta values in the last column.
    }
    \label{tab: main_results_llama}
\end{table*}

\section{Experiments}
\subsection{Exaperimental Setup}
\noindent \textbf{Model.} 
We select the models in the Qwen2.5-Instruct series \citep{qwen2.5} and Llama-3 series \citep{meta2024introducing} as our teachers and students. 
Specially, for the Qwen2.5 series, we select Qwen2.5-7B-Instruct as the teacher model, and Qwen2.5-1.5B-Instruct as the student model;
for the Llama-3 series, we select Llama3.1-8B-Instruct as the teacher model, and Llama3.2-1B-Intruct as the student model.

\noindent \textbf{Data.}
We evaluate the model on the instruction-following and math reasoning benchmarks.

For the instruction-following task, we choose the \texttt{databricks-dolly-15k} (Dolly) dataset processed by \citet{gu2024minillm} as the training set, which contains about 11k samples for training, 1k for validation, and 500 for testing. 

For the comprehensive evaluation, we select Self-Instruct (SelfInst), Vicuna-Evaluation (VicunaEval), Super Natural Instructions (S-NI), and Unnatural Instructions (UnNI) as the additional test sets, following \citet{gu2024minillm}.
For the math reasoning benchmarks, we consider the \texttt{Math} \citep{hendrycks2021measuring} as the train set.
For the evaluation, we select the Math-500 \citep{hendrycks2021measuring} and GSM8k \citep{cobbe2021training} as the test sets, where the MATH-500 has 500 cases uniformly sampled from Math, and the GSM8K has 1319 cases with different distribution from the train set.

\noindent \textbf{Training and Evaluation.}
For the instruction-following task, we first train the teacher and student models on the train set of Dolly, and then conduct the distillation on the same set, following \citet{gu2024minillm}.
During the evaluation, we measure the results of the instruct-following tasks by Rouge-L \citep{lin2004rouge}, which is suitable for large-scale instruction tuning evaluation \citep{wang2022super}.
For the math reasoning task, we directly conduct the distillation on the high-quality data sampled by the teacher, since the good performance of the pretrained models on the math tasks.
During the evaluation, we measure the math tasks by the accuracy with the off-the-shelf toolkit Qwen2.5-Math~\citep{yang2024qwen2}.
We train all our models on the open source OpenRLHF framework \citep{hu2024openrlhf}.
The hyperparameter setting and implementation details could be found in the the Appendix \ref{sec: app_hyper}.

\subsection{Baselines}
We conduct experiments with the following KD techniques on LLMs:
\begin{itemize} [leftmargin=*,topsep=0pt]
\setlength{\itemsep}{0pt}
\setlength{\parsep}{0pt}
\setlength{\parskip}{0pt}
    \item \textbf{SeqKD} \citep{kim2016sequence}. Directly learning the high-quality output data with SFT, which is a verifiable solution for LLMs distillation \citep{huang2024o1};
    \item \textbf{SKD} \citep{xu2024speculative}. SKD bridges the knowledge gaps between teacher and student in practical scenarios, by utilizing the teacher to refine the output of the student and take further optimization;
    % first leverages the teacher to refine the student model's output, and then optimizes the student model via SeqKD on the refined output.
    % 
    \item \textbf{FKL} \citep{hinton2015distilling}. The widely used standard KL divergence in knowledge distillation;
    \item \textbf{RKL} \citep{tu2020engine}. The reverse KL divergence swaps the two distributions in KL divergence;
    \item \textbf{f-Distill} \citep{wen2023f}. A framework that addresses KL divergence’s mode averaging and collapsing problems by minimizing a symmetric f-divergence;
    \item \textbf{s-FKL} \citep{ko2024distillm} The s-FKL skews the student distribution $q_{\theta}$ in FKL as $\lambda p + (1 - \lambda)q_{\theta}$.
    \item \textbf{aKL} \citep{wu2024rethinking} The adaptive fusion of FKL and RKL to focus the top and tail regions of the teacher's distribution, respectively.
\end{itemize}

\subsection{Main Results}
% We present the experimental results in Table \ref{tab: main_results}.
We first conduct experimental results by distilling the 7/8B models to the 1.5/1B models, and present results in Table~\ref{tab: main_results_qwen} and Table~\ref{tab: main_results_llama}, respectively.

Firstly, the current KD techniques could achieve better performance of distillation under the same teacher and training data, with our student$_{warmup}$.
As shown in Table~\ref{tab: main_results_qwen} and Table~\ref{tab: main_results_llama}, based on our warmup student$_{warmup}$, each tested KD method achieves the stronger student on both backbones, where the averaged improved score ranges from +0.36 to +0.89.
For instance, the improvement of FKL (+0.42/0.41) and RKL (+0.59/+53) with student$_{warmup}$ indicates the student model suffers from less mode-averaging and mode-collapsing problems during distillation.
In addition, the improvements of other KD techniques demonstrate that our student$_{warmup}$ is more suitable for distillation, and reducing the distribution mismatch in the early stage of distillation is beneficial.
Furthermore, with our student$_{warmup}$, the performance of various KD techniques in different settings of tasks and models suggests that our proposed Warmup-Distill could effectively reduce the interference of distribution mismatch with the distillation.

Secondly, for the more challenged math task, our student$_{warmup}$ could significantly improve the performance of distillation, \emph{e.g.}, the Qwen2.5-1.5B-Instruct model yields around +3.8\% accuracy on the Math-500 test and outperforms the vanilla SeqKD method by +1.9\% accuracy.
Besides, even for the GSM8K testset with different distribution from the train set of Math, our method could also assist the KD techniques to a further improvement with around +0.9/+1.8\% accuracy on the Qwen2.5-1.5B-Instruct and Llama3.2-1B-Instruct, respectively.

Overall, our proposed Warmup-Distill framework could effectively mitigate the distribution mismatch between the teacher and student models without extra human annotation, and further improve the performance of current KD techniques, no matter the black-box or white-box KD techniques.
% 

% \subsubsection{Loss Curves} \label{sec: loss}

\section{Discussion}

\begin{table}[t]
    \centering
    \resizebox{1.0\linewidth}{!}{
    \begin{tabular}{c|cccc | cccc}
        \toprule
        \textbf{Task} & \textbf{R} & \textbf{All} & \textbf{Len} & \textbf{Good} & \textbf{R} & \textbf{All} & \textbf{Len} & \textbf{Good} \\
        \midrule
        \multicolumn{9}{l}{Student: \texttt{Qwen2.5-1.5B-Instruct}; \quad Vanilla \emph{vs.} Warmup} \\
        \midrule
        \textbf{IF} & 29.89 & 19.78 & 83.77 & 73.05 & 50.13 & 36.27 & 80.99 & 84.23 \\
        \textbf{Math} & 21.36 & \;\,3.74 & 771.37 & 51.88 & 41.52 & 29.63 & 723.45 & 64.53 \\
        \midrule
        \multicolumn{9}{l}{Student: \texttt{Llama3.2-1B-Instruct}; \quad Vanilla \emph{vs.} Warmup}  \\
        \midrule
        \textbf{IF} & 23.99 & 12.05 & 92.04 & 66.64 & 32.76 & 23.81 & 101.62 & 74.33 \\
        \textbf{Math} & 16.20 & \;\,1.52 & 697.66 & 54.21 & 35.62 & 29.67  & 656.96 & 62.91 \\
        \bottomrule
    \end{tabular}
    }
    \caption{
    Data statistics of the output of student and student$_{warmup}$.
    IF and Math denote the Instruction-Following and Math tasks, respectively.
    \textbf{Len} denotes the averaged token numbers of student-generated sequences.
    \textbf{R}, \textbf{All} describe the averaged percentage of the tokens with the low probability margin, and the percentage of \textbf{R} equals 1.
    \textbf{Good} indicates the proportion of sequences improved by the teacher that are better than the original ones.
    }
    \label{tab: data_stat}
\vspace{-6pt}
\end{table}

\subsection{Data Distribution}
In this section, we present the details of the data distribution with teacher refinement in the warmup stage, and list the results in Table \ref{tab: data_stat}.

Generally, the R and All metrics reflect the degree of the distribution mismatch issue between the teacher and student, where the large value denotes the slighter issue.
As the Table \ref{tab: data_stat} shows, the warmup student could generate more tokens (higher R score) and more sequences (higher All score) within the distribution of the teacher model.
% As the Table \ref{tab: data_stat} shown, both student models have larger Position and Accepted values on the Instruct-Following task, compared to the Math task.
% 
This phenomenon suggests that our warmup student$_{warmup}$ suffers the less distribution mismatch issue in the early stage of distillation.

Besides, the Good metric also indicates the degree of the distribution mismatch issue.
In this metric, we leverage the Rouge-L and the Accuracy as the reward model to measure whether the teacher-refined sequences are better than the original ones, where the higher ratio of `Good' leads to the distribution of the student-generated sequences being more fitted with the teacher.
Similarly, the increased score of `Good' metric from the vanilla student to our warmup student takes a further evidence that  Warmup-Distill could effectively reduce the distribution mismatch issue between the teacher and student models in the warmup stage.
% Similar score distribution in the metric could further demonstrate that the larger knowledge gap exists in the Math task, and the more improvement in the benchmarks further demonstrates that Warmup-Distill could effectively reduce the knowledge gap before the distillation.

\begin{figure}[t!]
\begin{center}
    \resizebox{0.475\textwidth}{!}{
        \includegraphics[width=1\textwidth]{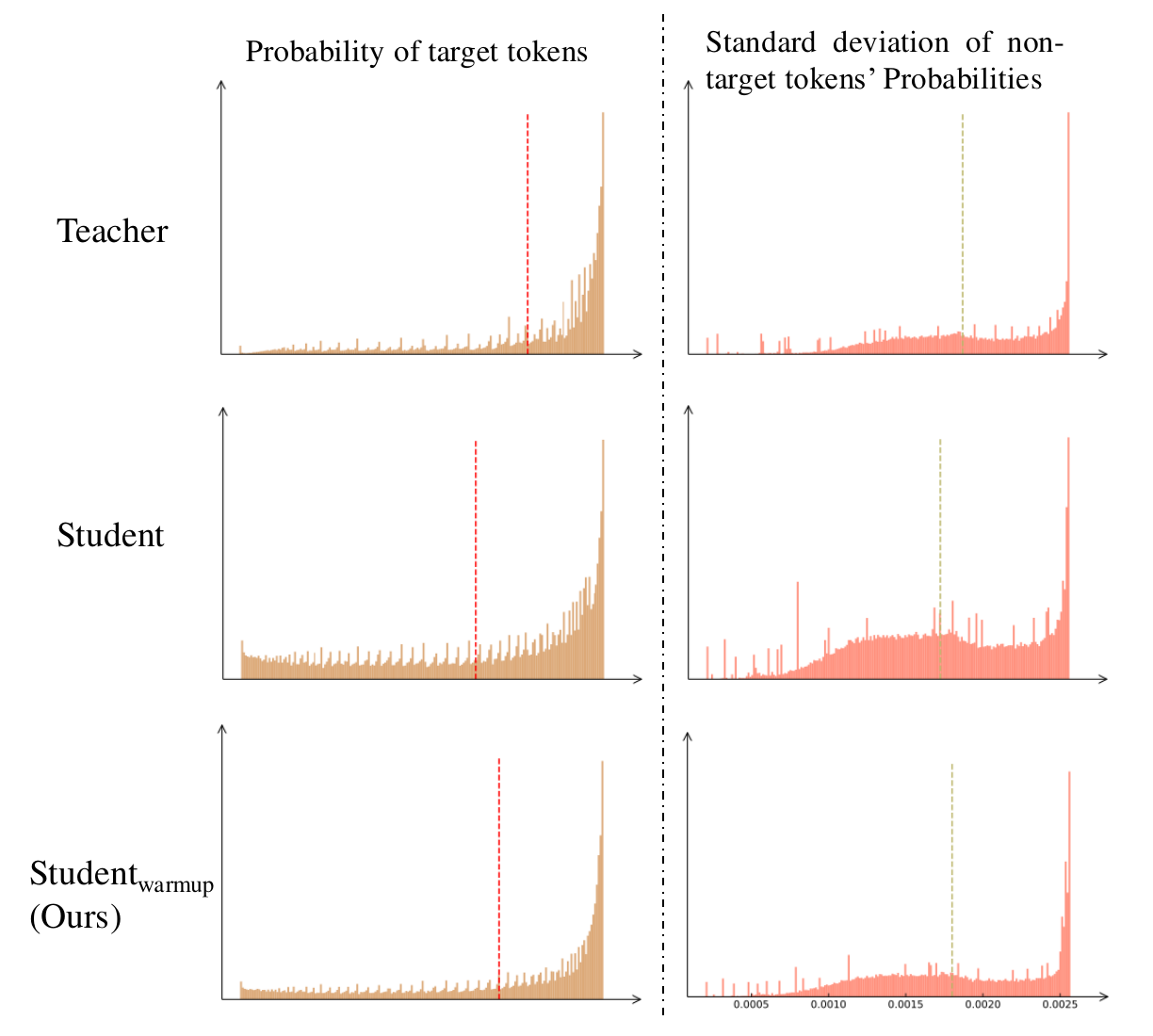}
    } 
    \caption{
    Probability distributions of the target tokens and the standard deviation of non-target tokens's probabilities to samples in the train set of the Math task.
    Rows show the teacher, vanilla student, and our student$_{warmup}$ models.
    Dashed lines denote the mean value.
    } 
    \label{fig: distribution}  
\vspace{-6pt}
\end{center} 
\end{figure}

\subsection{Similar Distribution}
To give a more intuitive representation that Warmup-Distill could mitigate the distribution mismatch issue in the early stage of distillation, we visualize the distribution of three models (Teacher, Student, Student$_{warmup}$) on the train set of Math based on Qwen series.

Following \citet{li2022asymmetric}, we compare the distribution of the probability of target tokens and the standard deviation of non-target tokens' probabilities, and plot the histograms in Figure~\ref{fig: distribution}.
% 
% Obviously, our student$_{warmup}$ tends to have a much more similar distribution of the logits of target tokens as the one of teacher (Left in Figure \ref{fig: distribution}), compared to the vanilla student.
% Obviously, our student$_{warmup}$ could spr
Obviously, our student$_{warmup}$ spreads the larger probability to the target tokens following the teacher model (Left in Figure \ref{fig: distribution}), compared to the vanilla student.
Besides, the more sharpened distribution of standard deviation of non-target tokens' probabilities (Right in Figure \ref{fig: distribution}) also indicates that the distribution of our student$_{warmup}$ is closer to the teacher's sharpened distribution.
In summary, the visualized distribution plot that our proposed Warmup-Distill could prepare a student$_{warmup}$ which has a more similar distribution with the teacher, and then suffers less interference of distribution mismatch during distillation.
% the more similar distribution between the student$_{warmup}$ and teacher models could further demonstrate that our proposed Warmup-Distill could effectively reduce the distribution mismatch issue and prepare a student more suitable for distillation.

\begin{figure}[t!]
\begin{center}
    % \scalebox{0.475}{
    \resizebox{0.475\textwidth}{!}{
        \includegraphics[width=1\textwidth]{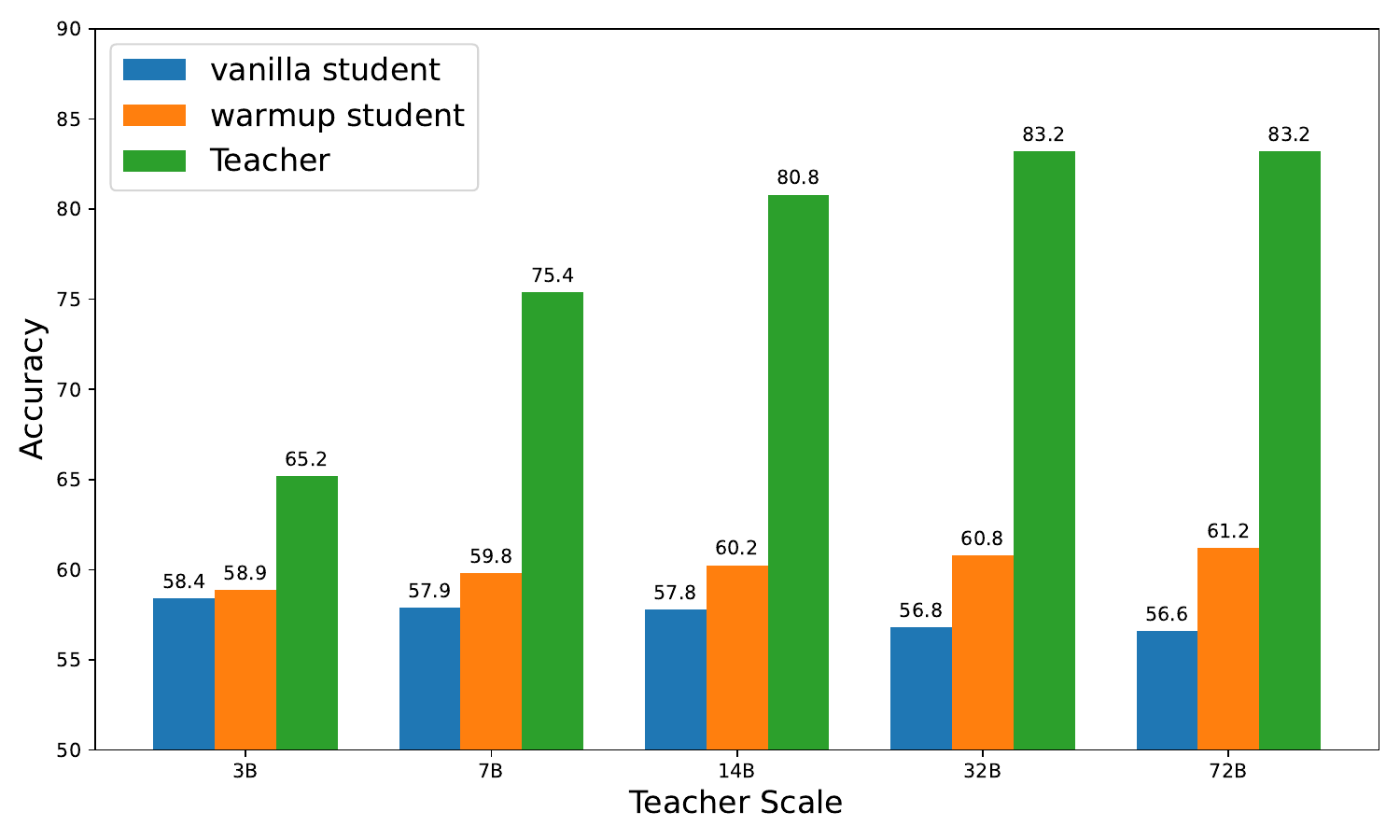}
    } 
    \caption{
        The scaling of teachers is based on the Qwen2.5 series and evaluated on MATH-500 set.
    } 
    \label{fig: large_teachers}  
\vspace{-6pt}
\end{center} 
\end{figure}
\subsection{Distillation with Larger Teacher}
We further conduct experiments to evaluate our proposed methods on the different teacher scales in the Qwen2.5 series using SeqKD techniques on the more challenging Math task.

In practice, the stronger teacher does not necessarily lead to a proportionally stronger student model, \emph{i.e.}, the Capacity Mismatch issue \citep{li2022asymmetric, qiu2022better, niu2024efficient, cui2024sinkhorn}.
As shown in Figure~\ref{fig: large_teachers}, with the scale of teachers increasing, the KD technique with the vanilla student cannot ensure the better student with the stronger teacher, which is limited by the Capacity Mismatch issue.
However, with our warmup student$_{warmup}$, the larger teacher could yield the stronger student, with the same KD technique, which displays the importance of reducing the distribution mismatch before distillation.
% 
% Besides, our proposed Warmup-Distill could take a further improvement for the stronger student model, \emph{e.g.}, DeepSeek-R1-Distill-Qwen-1.5B, as stated in Appendix \ref{sec: strong_student}.

% \begin{table}[t]
% \centering
% \begin{tabular}{lc}
% \hline
% \textbf{Command} & \textbf{Output}\\
% \hline
% \verb|{\"a}| & {\"a} \\
% \verb|{\^e}| & {\^e} \\
% \verb|{\`i}| & {\`i} \\ 
% \verb|{\.I}| & {\.I} \\ 
% \verb|{\o}| & {\o} \\
% \verb|{\'u}| & {\'u}  \\ 
% \verb|{\aa}| & {\aa}  \\\hline
% \end{tabular}
% \begin{tabular}{lc}
% \hline
% \textbf{Command} & \textbf{Output}\\
% \hline
% \verb|{\c c}| & {\c c} \\ 
% \verb|{\u g}| & {\u g} \\ 
% \verb|{\l}| & {\l} \\ 
% \verb|{\~n}| & {\~n} \\ 
% \verb|{\H o}| & {\H o} \\ 
% \verb|{\v r}| & {\v r} \\ 
% \verb|{\ss}| & {\ss} \\
% \hline
% \end{tabular}
% \caption{Example commands for accented characters, to be used in, \emph{e.g.}, Bib\TeX{} entries.}
% \label{tab:accents}
% \end{table}
\begin{table}[t]
    \centering
    \resizebox{0.9\linewidth}{!}{
    \begin{tabular}{c|c c c c c}
        \toprule
        \textbf{Sampling N} & 0 & 1 & 4 & \,8$^*$ & 12 \\
        \midrule
         \textbf{IF} & 29.07 & 29.29 & 29.45 & 29.76 & \textbf{29.84}\\
         \textbf{Math} & 66.85 & 67.11 & 67.73 & 68.05 & \textbf{68.17}\\
        \bottomrule
    \end{tabular}
    }
    \caption{
        Results of distillation based on Qwen2.5-1.5B-Instruction via SeqKD.
        Sampling N denotes the number of sampling sequences for each case, which is ranges from 1 to 12 in this section, and 0 denotes the vanilla student without the Warmup stage, 8 is the setting in our other experiments.
        Digital results of Instruct Following denote the averaged Rouge-L scores over the 5 test sets of instruct-following tasks, and results of Math Reasoning denote the averaged accuracy over the 2 test sets of math tasks in prior sections.
    }
    \label{tab: sampling_n}
\end{table}

\subsection{Hyperparameter} 
In this section, we investigate the impact of the number of sampling sequences for each case on our warmup student$_{warmup}$ model.

% and list the results in Table \ref{tab: sampling_n}.
Results are displayed in Table \ref{tab: sampling_n}, the performance of the student is generally increased from the small sampling number to the large one, since the more concrete distribution within the student model could be captured and aligned.
Compared to the selection of 8, the performance of the warmup student with sampling 12 sequences for each is not much better, with more expensive computational cost.
In other words, sampling 8 sequences could capture the most distribution within the student model could be captured and could be the optimal choice with the trade-off between performance and cost.

\begin{table}[t]
    \centering
    \resizebox{0.78\linewidth}{!}{
    \begin{tabular}{c|c c c c c}
        \toprule
        \textbf{Methods} & SFT & \,DPO$^*$ & Hinge & SimPO \\
        \midrule
         \textbf{IF} & 29.49 & 29.76 & 29.77 & \textbf{29.82} \\
         \textbf{Math} & 67.32 & 68.05 & \textbf{68.11} & 67.99 \\
        \bottomrule
    \end{tabular}
    }
    \caption{
        Results of distillation based on Qwen2.5-1.5B-Instruction with various methods to align the internal knowledge of the student to the teacher.
        DPO is our selection in other experiments.
    }
    \label{tab: sft_vs_dpo}
\vspace{-6pt}
\end{table}

\subsection{DPO vs. SFT for Alignment}
Further, we investigate which method is more suitable to align the distribution of the student to the teacher.

We mainly compare the SFT and three variants of DPO, \emph{i.e.}, vanilla DPO \citep{rafailov2024direct}, Hinge \citep{zhao2023slic}, and Simpo \citep{meng2024simpo}.
Results in Table \ref{tab: sft_vs_dpo} display that three variants of DPO perform better than SFT, whereas the performance among these three variants of DPO is similar.
Hence, we could directly adopt the DPO technique to optimize the student model in the warmup stage, where the variant of DPO is not matter.

% \subsection{Better student}
% \subsection{DPO varient}

% \subsection{Case Study}

\section{Related Work}
\subsection{Knowledge Distillation}
Knowledge distillation (KD) aims to compress the large model (Teacher) to a smaller one (Student) \citep{buciluǎ2006model, hinton2015distilling}. 
Consequently, SeqKD \citep{kim2016sequence} uses the teacher’s decoding sequences as the training data of the student and directly optimizes the cross-entropy loss on the one-hot target.
However, the distribution mismatch issue hinders the performance of distillation, since the student model struggles to learn the correct mode from the distribution of the teacher.
To solve this issue, previous studies investigate to adjust the calculated function towards the distribution between both models.
\citet{wen2023f} propose the symmetric f-divergence to avoid this issue during distillation;
Further, \citet{gu2024minillm} and \citet{ko2024distillm} mitigate this issue via adjusting the target distribution by combining the objective of both models.
Besides, \citet{wu2024rethinking} optimize the target distribution by adaptively fusing the values of FKL and RKL to focus different region of distribution of the teacher.
Moreover, \citet{xu2024speculative} bridges the knowledge gap between the teacher and student by optimizing the student with the teacher-refined output of the student.
However, the above methods mainly optimize this issue during distillation, ignoring the negative impact at the early stage of distillation.
% Besides, researchers investigate the various distilled objectives, \emph{e.g.}, the output of teachers' layers \citep{sun2019patient}, the attention and hidden state alignment \citep{sanh2019distilbert, jiao2019tinybert}.
% % 
% Recent research investigates the knowledge distillation of the LLMs to refine the process of knowledge transfer.
% % 
% \citet{gu2024minillm} propose a PPO-based framework with hybrid sampling to balance teacher guidance and student exploration. 
% % 
% Besides, \citet{liang2023less} propose the task-aware, layer-aware distillation to transfer the intermediate knowledge in the layers of the teacher.
% % % 
% Similarly, \citet{zhang2024dual} focuses on the distillation on the last layer's output hidden states of teacher and student models.

\subsection{Preference Optimization}
The Reinforcement Learning from Human Feedback (RLHF;~\citet{christiano2017deep}) framework initially trains a reward model on preference data and fine-tunes language models (LMs).
DPO \citep{rafailov2024direct} bypasses explicit reward modeling by optimizing a pairwise logistic loss directly on preferences. 
SLiC \cite{zhao2023slic} adopts pairwise hinge losses, and SimPO \citep{meng2024simpo} utilizes the average log probability of the sequence as the implicit reward to better align the reward with model generation.
% 
% In this paper, we leverage 
% while RSO \citep{liu2023statistical} enhances DPO via rejection sampling to mitigate distribution drift.
% 

\section{Conclusion}
In this paper, we focus on the Distribution Mismatch issue during knowledge distillation.
Different from the previous works, we propose the Warmup-Distill framework to align the distribution of the student to the teacher before distillation.
Experimental results demonstrate that our warmup student$_{warmup}$ could significantly adjust the distribution of the student to be similar with the one of the teacher, and prove our proposed Warmup-Distill method could effectively assist the current KD techniques to achieve better performance.

\section*{Limitations}
In practice, a few declines in the KD techniques with our warmup student$_{warmup}$, which may be due to the alignment tax during DPO \citep{lin2024mitigating, lu2024online}.
Since this issue is beyond the scope of this research, we leave the refinement to future work.
% 
% PPO
% ACL 2023 requires all submissions to have a section titled ``Limitations'', for discussing the limitations of the paper as a complement to the discussion of strengths in the main text. This section should occur after the conclusion, but before the references. It will not count towards the page limit.
% The discussion of limitations is mandatory. Papers without a limitation section will be desk-rejected without review.

% While we are open to different types of limitations, just mentioning that a set of results have been shown for English only probably does not reflect what we expect. 
% Mentioning that the method works mostly for languages with limited morphology, like English, is a much better alternative.
% In addition, limitations such as low scalability to long text, the requirement of large GPU resources, or other things that inspire crucial further investigation are welcome.

% \section*{Ethics Statement}
% Scientific work published at ACL 2023 must comply with the ACL Ethics Policy.\footnote{\url{https://www.aclweb.org/portal/content/acl-code-ethics}} We encourage all authors to include an explicit ethics statement on the broader impact of the work, or other ethical considerations after the conclusion but before the references. The ethics statement will not count toward the page limit (8 pages for long, 4 pages for short papers).

% \section*{Acknowledgements}

\bibliography{anthology,custom}
\bibliographystyle{acl_natbib}

\appendix

\section{Implementation Details} \label{sec: app_hyper}
% \label{sec:appendix}
In this section, we provide the details of the implementation of our experiments.

\subsection{Data}
\noindent \textbf{Instruction Following}.
We select the \texttt{databricks-dolly-15k} (Dolly) dataset processed by \citet{gu2024minillm} as the training set, which contains about 11k samples for training, 1k for validation, and 500 for testing. 
For all test sets, Dolly contains 500 samples, Self-Instruction \citep{wang2022self} contains 242 samples, Vicuna-Evaluation \citep{vicuna2023} contains 80 samples, Super-Natural Instructions \citep{wang2022super} contains 1694 samples with response lengths in $[11, +\infty]$, and Unnatural Instructions \citep{honovich2022unnatural} contains 10000 samples
with response lengths in $[11, +\infty]$.

\noindent \textbf{Math Reasoning}.
We leverage the teacher models to generate their output as the training set.
In the original training set of Math, 7500 cases are included, and we sample 8 sequences for each case and choose the correct output from the teacher as the training set for the student.
For the test set of MATH-500 and GSM8K, the prior set includes 500 cases with 5 different difficult levels, and the latter has 1319 cases.

\subsection{Training}
\noindent \textbf{Instruction Following}.
To initialize the teacher and student models, we employ the full-finetuning to LLMs.
For the Qwen2.5 models, we finetune the teacher with 3 epochs, and 1 epoch for the student; and 10 epochs for the teacher of Llama model, and 3 epochs for the student.
In the hyperparameters, we finetune all models by setting the learning rate to 2e-5, batch size to 64, max length of sequence to 1k, with the Cosine LR scheduler.
For the distillation, we continue to utilize the above hyperparameters, and set the coefficient to 0.5 for different white-box KD methods, and temperature $\tau$ to 2.0:
\begin{equation}
    \mathcal{L} = 0.5 \times \mathcal{L}_{ce} + 0.5 \times \mathcal{L}_{kd}.
\end{equation}
For the black-box KD methods, based on the teacher-generated data or teacher-refined output of student filtered by Rouge-L, we optimize the student with $\mathcal{L}_{ce}$.

\noindent \textbf{Math Reasoning}.
We directly use the LLMs as the teacher or student model.
During distillation, we optimize the student model on the teacher-generated sequences for 1 epoch, with 2e-6 as the learning rate.
And we set the batch size to 16, the length of sequence to 4K, with the Cosine LR scheduler.

\begin{table}[t]
    \centering
    \resizebox{0.78\linewidth}{!}{
    \begin{tabular}{c|c c c c c}
        \toprule
        \textbf{$\eta$} & default & 2 & \,4$^*$ & 8 & 16 \\
        \midrule
         \textbf{IF} & 29.07 & 28.46 & \textbf{29.76} & 29.69 & 29.55 \\
         \textbf{Math} & 66.85 & 63.93 & \textbf{68.05} & 68.02 & 67.64 \\
        \bottomrule
    \end{tabular}
    }
    \caption{
          Results of distillation based on Qwen2.5-1.5B-Instruction via SeqKD.
        $\eta$ denotes the value of the threshold, which is ranges from 2 to 16 in this section, and default denotes the vanilla student without the Warmup stage, 4 is the setting in our other experiments.
    }
    \label{tab: thres}
\end{table}

\subsection{Evaluation}
For the evaluation of instruction-following benchmarks, we use the sampling strategy to decode the responses from all models. 
For decoding, we set both the decoding temperature and top\_p to 1.0, with 1024 max new tokens, and we report the Rouge-L scores for each benchmark.
For the evaluation of math reasoning benchmarks, we also use the sampling strategy to decode the response with 4K max new tokens, and measure the output with the off-the-shelf toolkit Qwen2.5-Math~\citep{yang2024qwen2}. 
Since the Long generation in these tasks, we adopt the vLLM\footnote{version: 0.6.4.post1} \citep{kwon2023efficient} as the inference framework to accelerate the generation.

\subsection{Implementation}
We train all our models on the open source OpenRLHF framework \citep{hu2024openrlhf} on eight H20 GPUs, and set the seeds in [0, 42, 123] to repeat the training and report the a
aged score.
For the DPO model training in the distribution alignment, we set the learning rate to 1e-7, while the other settings are the same as the above settings.
To simplify the collection of the pairwise data with mismatched distribution, we convert the score in Equation from the probability to the corresponding rank, where only one model is loaded, and set the threshold parameter $\eta$ to 4.
To the selection of $\eta$, we observe cases in the early experiments, where the smaller $\eta$ will filter out amount of common words, such as `the', `we', and so on.
Besides, the too larger $\eta$ will miss the tokens with the mismatched distribution issue.

From the perspective of the performance of final distillation, we list our exploration results in Table \ref{tab: thres} for reference.
The smaller $\eta$ will eliminate the original knowledge within the student model and provide a negative impact on subsequence distillation, while the large $\eta$ will ignore mismatched tokens and could not lead to optimal distillation performance.
\end{document}